\documentclass[11pt,a4paper]{article}
\usepackage[hyperref]{acl2021}
\usepackage{times}
\usepackage{latexsym}

\usepackage{adjustbox}
\usepackage{amssymb,amsmath}
\usepackage{multirow}

\usepackage{microtype}

\aclfinalcopy 
\def\aclpaperid{3774} 

\title{Discrete Cosine Transform as Universal Sentence Encoder}

\author{Nada Almarwani $^{\textbf{1,2}}$ \and Mona Diab $^{\textbf{1,3}}$\\
$^1$ Dep. of Computer Science, The George Washington University\\
$^2$ Dep. of Computer Science, College of Computer Science and Engineering, Taibah University\\
$^3$ Facebook AI Research \\
{\tt nmarwani@taibah.edu.sa, mdiab@fb.com}
}
\date{}

\begin{document}
\maketitle
\begin{abstract}
Modern sentence encoders are used to generate dense vector representations that capture the underlying linguistic characteristics for a sequence of words, including phrases, sentences, or paragraphs. These kinds of representations are ideal for training a classifier for an end task such as sentiment analysis, question answering and text classification. Different models have been proposed to efficiently generate general purpose sentence representations to be used in pretraining protocols. While averaging is the most commonly used efficient sentence encoder, Discrete Cosine Transform (DCT) was recently proposed as an alternative that captures the underlying syntactic characteristics of a given text without compromising practical efficiency compared to averaging. However, as with most other sentence encoders, the DCT sentence encoder was only evaluated in English. To this end, we utilize DCT encoder to generate universal sentence representation for different languages such as German, French, Spanish and Russian. The experimental results clearly show the superior effectiveness of DCT encoding in which consistent performance improvements are achieved over strong baselines on multiple standardized datasets. 
\end{abstract}

\section{Introduction}
Recently, a number of sentence encoding representations have been developed to accommodate the need of sentence-level understanding; some of these models are discussed in \cite{hill-etal-2016-learning-distributed,logeswaran2018efficient,conneau-etal-2017-supervised}, yet most of these representations have focused on English only. 

To generate sentence representations in different languages, the most obvious solution is to train monolingual sentence encoders for each language. However, training a heavily parameterized monolingual sentence encoder for every language is inefficient and computationally expensive, let alone the impact on the environment. Thus, utilizing a non-parameterized model with ready-to-use word embeddings is an efficient alternative to generate sentence representations in various languages. 

A number of non-parameterized models have been proposed to derive sentence representations from pre-trained word embeddings \cite{ruckle2018concatenated,yang-etal-2019-parameter,kayal-tsatsaronis-2019-eigensent}. However, most of these models, including averaging, disregard structure information, which is an important aspect of any given language. Recently, \newcite{almarwani-etal-2019-efficient} proposed a structure-sensitive sentence encoder, which utilizes Discrete Cosine Transform (DCT) as an efficient alternative to averaging. The authors show that this approach is versatile and scalable because it relies only on word embeddings, which can be easily obtained from large unlabeled data. Hence, in principle, this approach can be adapted to different languages. Furthermore, having an efficient, ready-to-use language-independent sentence encoder can enable knowledge transfer between different languages in cross-lingual settings, empowering the development of efficient and performant NLP models for low-resource languages. 

In this paper, we empirically investigate the generality of DCT representations across languages as both a single language model and a cross-lingual model in order to assess the effectiveness of DCT across different languages.

\section{DCT as sentence Encoder}
In signal processing domain DCT is used to decompose signal into component frequencies revealing dynamics that make up the signal and transitions within \cite{shu2017study}. Recently, DCT has been adopted as a way to compress textual information \cite{kayal-tsatsaronis-2019-eigensent,almarwani-etal-2019-efficient}. A key observation in NLP is that word vectors obey laws of algebra King – Man + Woman = (approx.) Queen \cite{mikolov2013distributed}. Thus, given word embeddings, cast a sentence as a multidimensional signal over time, in which DCT is used to summarize the general feature patterns in word sequences and compress them into fixed-length vectors \cite{kayal-tsatsaronis-2019-eigensent,almarwani-etal-2019-efficient}. 

Mathematically, DCT is an invertible function that maps an input sequence of N real numbers to the coefficients of N orthogonal cosine basis functions of increasing frequencies \cite{ahmed1974discrete}. The DCT components are arranged in order of significance. The first coefficient (c[0]) represents the sum of the input sequence normalized by the square length, which is proportional to the average of the sequence \cite{ahmed1974discrete}. The lower-order coefficients represent lower signal frequencies which correspond to the overall patterns in the sequence. For example, DCT is used for compression by preserving only the coefficients with large magnitudes. These coefficients can be used to reconstruct the original sequence exactly using the inverse transform \cite{watson1994image}. 
 
In NLP, \newcite{kayal-tsatsaronis-2019-eigensent} applied DCT at the word level to reduce the dimensionality of the embeddings size, while \newcite{almarwani-etal-2019-efficient} applied it along the sentence length as a way to compress each feature in the embedding space independently. In both implementations, the top coefficients are concatenated to generate the final representation for a sentence. As shown in \cite{almarwani-etal-2019-efficient}, applying DCT along the features in the embeddings space renders representations that yield better results. Also, \newcite{zhu-de-melo-2020-sentence} noted that similar to vector averaging the DCT model proposed by \cite{almarwani-etal-2019-efficient} yields better overall performance compared to more complex encoders, thus, in this work, we adopt their implementation to extract sentence-level representations. 

Specifically, given a sentence matrix $N \times d$, a sequence of DCT coefficients $c[0], c[1], ..., c[K]$ are calculated by applying the DCT type II along the $d$-dimensional word embeddings, where \(c[K]=\sqrt{\frac{2}{N}}\sum_{n=0}^{N-1}v_n \cos{\frac{\pi}{N}(n+\frac{1}{2})K}\) \cite{shao2008type}. Finally, a fixed-length sentence vector of size $Kd$ is generated by concatenating the first $K$ DCT coefficients, which we refer to as $c[0:K]$.\footnote{Unlike \cite{almarwani-etal-2019-efficient}, we note no further improvements with larger coefficients, thus, we only report the results of \(1 \leq K \leq 4\).}
\begin{figure*}[!hbt]
\centering\includegraphics[width=\textwidth]{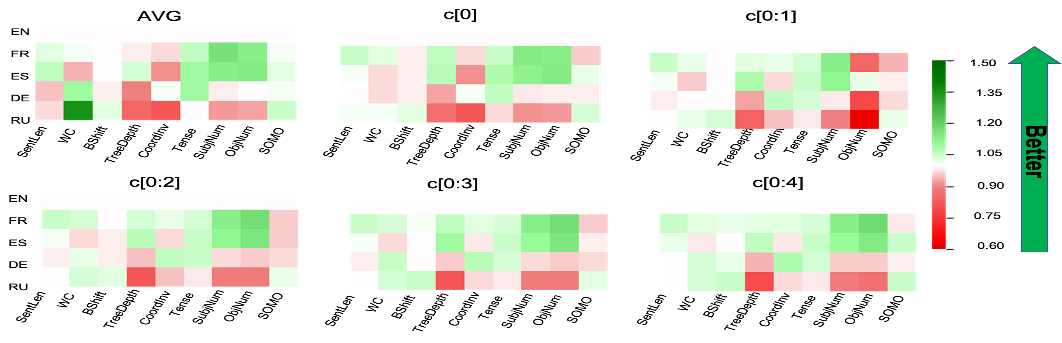}
\caption{Results of the probing tasks comparing XX languages performance relative to English. White indicates a value of 1, demonstrating parity in performance with English. Red indicates better English performance while green indicates better XX Lang results.}
\label{X-PROBE-relative-Results}
\end{figure*}
 \begin{table}[!t]
\centering
\setlength\tabcolsep{4pt}
\begin{tabular}{|l|l|}
\hline
\textbf{Task}&\textbf{Description}\\\hline
SentLen&Length prediction\\\hline
WC&Word Content analysis\\\hline
BShift&Word order analysis\\\hline
TreeDepth&Tree depth prediction\\\hline
Tense&Verb tense prediction\\\hline
CoordInv&Coordination Inversion\\\hline
SubjNum&Subject number prediction\\\hline
ObjNum&Object number prediction\\\hline
SOMO&Semantic odd man out\\\hline

\end{tabular}
\caption{Probing Tasks as described in \cite{conneau-etal-2018-cram,ravishankar-etal-2019-probing}.}
\label{Summary}
\end{table}
\section{Multi-lingual DCT Embeddings }
\label{multi-EMBED}

\subsection{Experimental Setups and Results}
In our study, DCT is used to learn a separate encoder for each language from existing monolingual word embeddings. To evaluate DCT embeddings across different languages, we used the probing benchmark provided by \newcite{ravishankar-etal-2019-probing}, which includes a set of multi-lingual probing datasets.\footnote{Refer to \cite{conneau-etal-2018-cram} and \cite{ravishankar-etal-2019-probing} for more details about the probing tasks.} The benchmark covers five languages: English, French, German, Spanish and Russian, derived from Wikipedia. The task set comprises 9 probing tasks, summarized in Table \ref{Summary}, that address varieties of linguistic properties including surface, syntactic, and semantic information \cite{conneau-etal-2018-cram,ravishankar-etal-2019-probing}. \newcite{ravishankar-etal-2019-probing} used the datasets to evaluate different sentence encoders trained by mapping sentence representations to English. Unlike \newcite{ravishankar-etal-2019-probing}, we use the datasets to evaluate DCT embeddings for each language independently. As a baseline, in addition to the DCT embeddings, we use vector averaging to extract sentence representations from the pre-trained embeddings. 

For model evaluations, we utilize the SentEval framework introduced in \cite{conneau-kiela-2018-senteval}. In all experiments, we use a single-layer MLP on top of DCT sentence embeddings with the following parameters: kfold=10, batch\_size=128, nhid=50, optim=adam, tenacity=5, epoch\_size=4. For the word embeddings, we relied on the publicly available pre-trained FastText embeddings introduced in \cite{grave-etal-2018-learning}.\footnote{ Available at: https://fasttext.cc.}

\paragraph{Results:} Figure \ref{X-PROBE-relative-Results} shows a heat-map reflecting the probing results of the different languages relative to English. Overall, French (FR) seems to be the closest to English (EN) followed by Spanish (ES) then German (DE) and then finally Russian (RU) across the various DCT coefficients. Higher coefficients reflect majority better performance across most tasks for FR, ES and DE. We see the most variation with worse results than English on the syntactic tasks of TreeDepth, CoordInv, Tense, SubjNum and ObjNum for RU. SOMO stands out for RU where it outperforms EN. The variation in Russian might be due to the nature of RU being a more complex language that is morphologically rich with flexible word order \cite{toldova2015evaluation}.

In terms of the performance per number of DCT coefficients, we observe consistent performance gain across different languages that is similar to the English result trends. Specifically, for the surface level tasks, among the DCT models the $c[0]$ model significantly outperforms the $AVG$ with an increase of $\sim$30 percentage points in all languages. The surface level tasks (SentLen and WC) show the most notable variance in performance, in which the highest results are obtained using the $c[0]$ model. However, the performance decreases in all languages when K is increased. On the other hand, for all languages, we observe a positive effect on the model's performance with larger K in both the syntactic and semantic tasks. The complete numerical results are presented in the Appendix in Table \ref{X-PROBE-Full-Results}.

\section{Cross-lingual Mapping based on DCT Encoding}
\label{Cross-EMBED}
\subsection{Approach}
\newcite{aldarmaki-diab-2019-context}
proposed sentence-level transformation approaches to learn context-aware representations for cross-lingual mappings. While the word-level cross-lingual transformations utilize an aligned dictionary of word embeddings to learn the mapping, the sentence-level transformations utilize a large dictionary of parallel sentence embeddings. Since sentences provide contexts that are useful for disambiguation for the individual word's specific meaning, sentence-level mapping yields a better cross-lingual representation compared to word-level mappings.

A simple model like sentence averaging can be used to learn transformations between two languages as shown in \cite{aldarmaki-diab-2019-context}. However, the resulting vectors fail to capture structural information such as word order, which may result in poor cross-lingual alignment. Therefore, guided by the results shown in \cite{aldarmaki-diab-2019-context}, we further utilize DCT to construct sentence representations for the sentence-level cross-lingual modeling. 
\subsection{Experiments Setups and Results} 
For model evaluation, we use the same cross-lingual evaluation framework introduced in \cite{aldarmaki-diab-2019-context}. Intuitively, sentences tend to be clustered with their translations when their vectors exist in a well-aligned cross-lingual space. Thus, in this framework, cross-lingual mapping approaches are evaluated using sentence translation retrieval by calculating the accuracy of correct sentence retrieval. Formally, the cosine similarity is used to find the nearest neighbor for a given source sentence from the target side of the parallel corpus. 

\subsection{Evaluation Datasets and Results}
To demonstrate the efficacy of cross-lingual mapping using the sentence-level representation generated by DCT models, similarly to \newcite{aldarmaki-diab-2019-context}, we used the WMT'13 data set that includes EN, ES and DE languages \cite{bojar-etal-2013-findings}. We further used five language pairs from the WMT'17 translation task to evaluate the effectiveness of DCT-based embeddings. Specifically, we used a sample of 1 million parallel sentences from WMT’13 common-crawl data; this subset is the same one used in \cite{aldarmaki-diab-2019-context}.\footnote{Evaluation scripts and WMT'13 dataset as described in \cite{aldarmaki-diab-2019-context} are available in https://github.com/h-aldarmaki/sent\_translation\_retrieval} To assess efficacy of the DCT models for the cross-lingual mapping, we reported the performances of the sentence translation retrieval task within the WMT’13 test set, which includes EN, ES, and DE as test languages \cite{bojar-etal-2013-findings}. Specifically, we first used the 1M parallel sentences for the alignment between source languages (ES and DE) to a target language (EN) independently. We evaluated the translation retrieval performance in all language directions, from source languages to English: ES-EN and DE-EN, as well as between the sources languages: ES-DE. 

Similarly, we conduct a series of experiments on 5 different language pairs from the WMT'17 translation task, which includes DE, Latvian (LV), Finnish (FI), Czech (CS), and Russian (RU), each of which is associated with an English translation \cite{zhang-etal-2018-accelerating}.\footnote{The pre-processed version of the WMT'17 dataset was used. For more information refer to \cite{zhang-etal-2018-accelerating}.} For each language pair, we sampled 1M parallel sentences from their training corpus for the cross-lingual alignment between each source language and EN. Also, we used the test set available for each language pair to evaluate the translation retrieval performances.

In our experiments, we evaluate the translation retrieval performance in all language directions using three type of word embeddings: 1- a publicly available pre-trained word embeddings in which we show the performance of DCT against averaging, which we refer to hereafter as out-of-domain embeddings as shown in Table \ref{ALL-Results}. 2- Also, we ran additional experiments in which we used a domain specific word embedding (that we trained on genre that is similar to the translation task) and 3-contextualized word embedding, which we refer to hereafter as in-domain embeddings as shown in Table \ref{In-domain-Embeddings}. 
\begin{table}[t]
\centering
\begin{adjustbox}{width=.46\textwidth}
 \begin{tabular}{|l|c|c|c|c|c|}
\hline
Lang pair&$AVG$&$c[0]$&$c[0:1]$&$c[0:2]$&$c[0:3]$\\\noalign{\hrule height 1pt}
\multicolumn{6}{|l|}{\textbf{Lang}$\rightarrow$\textbf{EN}}\\\hline
 ES$ \rightarrow $EN&65.67&64.87&71.26&\textbf{71.80}&70.13\\\hline
 DE$ \rightarrow $EN&51.80&50.30&57.23&\textbf{58.13}&56.57\\\hline
 RU$ \rightarrow $EN&45.22&52.75&61.91&\textbf{64.35}&63.33\\\hline
CS$ \rightarrow $EN&41.87&42.50&52.89&54.99&\textbf{55.05}\\\hline
FI$ \rightarrow $EN&40.46&42.00&47.57&\textbf{47.80}&46.16\\\hline
LV$ \rightarrow $EN&21.26&40.13&51.42&56.37&\textbf{60.16}\\\noalign{\hrule height 1pt}
\multicolumn{6}{|l|}{\textbf{\textbf{EN}$\rightarrow$\textbf{Lang}}}\\\hline
 EN$ \rightarrow $ES&69.97&69.50&73.73&\textbf{73.87}&71.73\\\hline
 EN$ \rightarrow $DE&67.50&66.23&\textbf{69.27}&68.70&65.83\\\hline
 EN$ \rightarrow $RU&38.09&44.29&54.73&59.51&\textbf{60.94}\\\hline
 EN$ \rightarrow $CS&39.73&40.40&50.99&54.00&\textbf{54.12}\\\hline
 EN$ \rightarrow $FI&39.34&42.52&51.67&\textbf{52.59}&51.74\\\hline
 EN$ \rightarrow $LV&15.83&33.55&47.08&53.22&\textbf{55.72}\\ \noalign{\hrule height 1pt}
\multicolumn{6}{|l|}{\textbf{\textbf{Lang1}$\rightarrow$\textbf{Lang2}}}\\\hline
 DE$ \rightarrow $ES&43.80&42.20&49.50&\textbf{51.20}&51.17\\\hline
 ES$ \rightarrow $DE&57.67&56.46&\textbf{60.53}&59.83&57.87\\\hline
\end{tabular}
\end{adjustbox}
\caption{Sentence translation retrieval accuracy based on out of domain pre-trained Fasttext embeddings. Arrows indicate the direction, with English ($EN$), Spanish ($ES$), German ($DE$), Russian ($RU$), Czech ($CS$), Finnish ($FI$) , Turkish ($TR$), and Latvian ($LV$).}
\label{ALL-Results}
\end{table}

\textbf{Out-of-domain embeddings:} For all language pairs, DCT-based models outperform AVG and c[0] models in the sentence translation retrieval task. In the direction $\rightarrow EN$, while the c[0:2] model achieve the highest accuracy for ES, DE, RU, and FI languages, the c[0:3] model achieved the highest accuracy for CS and LV languages. Specifically, the c[0:2] model yields increases of $~$5.59\%-30\% in the direction from source languages (ES, DE, RU, and FI) to English compared to the AVG model. Also, while the c[0:3] model yielded an increase of $~$13\% gains over the baseline for CS, it provides the most notable increase of $~$38\% for LV.
For the opposite directions $EN \rightarrow source$, the DCT-based embeddings model also outperformed AVG and c[0] models. In particular, we observed accuracy gains of at least 3.81\% points using more coefficients in DCT-based models compared to the AVG and c[0] models for all languages.
A similar trend is observed in the zero-shot translation retrieval between the two non English languages ($ES$ and $DE$), in which DCT-based models outperform the AVG and c[0] models.

\begin{table}[!t]
\centering
\begin{adjustbox}{width=0.5\textwidth}
 \begin{tabular}{|l|l|c|c|c|c|c|}
\hline
Lang pair&Embed&$AVG$&$c[0]$&$c[0:1]$&$c[0:2]$&$c[0:3]$\\\noalign{\hrule height 1pt}
\multicolumn{7}{|l|}{\textbf{Lang}$\rightarrow$\textbf{EN}}\\\hline
\multirow{2}{*}{ ES$ \rightarrow $EN}&FT&82.97&82.40&\textbf{84.50}&83.97&82.90\\
&BERT&92.10&92.00&\colorbox{gray!40}{\textbf{93.23}}&93.13&92.20\\\hline
\multirow{2}{*}{ DE$ \rightarrow $EN}&FT&79.33&78.73&\textbf{81.87}&80.20&77.93\\
&BERT&89.76&89.66&\colorbox{gray!40}{\textbf{91.83}}&91.20&90.57\\\noalign{\hrule height 1pt}
\multicolumn{7}{|l|}{\textbf{EN}$\rightarrow$\textbf{Lang}}\\\hline
\multirow{2}{*}{ EN$ \rightarrow $ES}&FT&82.33&82.07&\textbf{85.47}&84.60&83.17\\
&BERT&93.63&93.66&\colorbox{gray!40}{\textbf{94.10}}&94.00&92.80\\\hline
\multirow{2}{*}{ EN$ \rightarrow $DE}&FT&74.73&74.50&\textbf{79.10}&78.70&76.90\\
&BERT&91.30&91.43&\colorbox{gray!40}{\textbf{91.90}}&91.53&90.30\\\noalign{\hrule height 1pt}
\multicolumn{7}{|l|}{\textbf{Lang1}$\rightarrow$\textbf{Lang2}}\\\hline
\multirow{2}{*}{ DE$ \rightarrow $ES}&FT&73.27&72.20&\textbf{77.43}&75.96&74.60\\
&BERT&87.80&87.57&90.23&\colorbox{gray!40}{\textbf{90.36}}&88.96\\\hline
\multirow{2}{*}{ ES$ \rightarrow $DE}&FT&68.90&68.07&\textbf{73.97}&73.10&72.43\\
&BERT&87.70&87.70&\colorbox{gray!40}{\textbf{89.67}}&89.50&88.53\\\hline
\end{tabular}
\end{adjustbox}
\caption{Accuracy using in-domain FastText (FT) and Contextualized mBERT embeddings. The best results for each row in \textbf{Bold} \& for each direction in \colorbox{gray!40}{\textbf{gray}}.}
\label{In-domain-Embeddings}
\end{table}

\textbf{In-domain embeddings:} To ensure comparability to state-of-the-art results, we further utilized in-domain FastText embeddings as those used in \cite{aldarmaki-diab-2019-context} as well as contextualized-based word embeddings. For the in-domain FastText embeddings, the FastText \cite{bojanowski-etal-2017-enriching} is utilized to generate word embeddings from 1 Billion Word benchmark \cite{DBLP:conf/interspeech/ChelbaMSGBKR14} for English, and equivalent subsets of about 400 million tokens from WMT'13 \cite{bojar-etal-2013-findings} news crawl data. For the contextualized-based embeddings, we utilized multilingual BERT (mBERT) introduced in \cite{devlin-etal-2019-bert} as contextual word embeddings, in which representations from the last BERT layer are taken as word embeddings. 
As shown in Table \ref{In-domain-Embeddings}, using in-domain word embeddings yields stronger results compared to the pre-trained embeddings we use in the previous experiments as illustrated in Table \ref{ALL-Results}. On the other hand, we observe additional improvements using mBERT as word embeddings on all models. Furthermore, increasing $K$ has positive effect on both embeddings, in which $c[0:1]$ demonstrate performance gains compared to other models in all language directions. This trend is clearly observed in the zero-shot performance between the non English languages.

\begin{table}
\centering
\begin{adjustbox}{width=0.45\textwidth}
\setlength\tabcolsep{4pt}
\begin{tabular}{|l|c|}
\hline 
\textbf{Model}&\textbf{ Average Accuracy}\\\hline
FastText (dict) [ALD2019]& 69.04\\\hline
ELMo (word) [ALD2019]&82.23\\\hline 
FastText (word) [ALD2019]&74.00\\\hline
FastText $AVG$ (sent) [ALD2019]& 76.92\\\hline
ELMo $AVG$ (sent) [ALD2019]&84.03\\\hline\noalign{\hrule height 1pt}
FastText $c[0]$ (sent)& 76.33\\\hline
FastText $c[0:1]$ (sent)& 80.39\\\hline
FastText $c[0:2]$ (sent)& 79.42\\\hline 
FastText $c[0:3]$ (sent)& 77.99\\\hline\noalign{\hrule height 1pt}
mBERT $AVG$ (sent)&90.38\\\hline
mBERT $c[0]$ (sent)&90.34\\\hline
mBERT $c[0:1]$ (sent)&\textbf{91.83}\\\hline
mBERT $c[0:2]$ (sent)&91.62\\\hline
mBERT $c[0:3]$ (sent)&90.56\\\hline
\end{tabular}
\end{adjustbox}
\caption{The average accuracy of various models across all language retrieval directions as reported in \cite{aldarmaki-diab-2019-context}, refer to as [ALD2019] in the table, along with the different DCT-based models in this work, in which (word) refers to word-level mapping, (sent) refers to sentence-level mapping, and (dict) refers to the baseline (using a static dictionary for mapping). \textbf{Bold} shows the best overall result.}
\label{AVG-ACC}
\end{table} 
 
Furthermore, as shown in Table \ref{AVG-ACC}, we obtained a state-of-the-art result using mBERT $c[0:1]$ with \textbf{91.83\%} average accuracy across all translation directions compared to the 84.03\% average accuracy of ELMo as reported in \cite{aldarmaki-diab-2019-context}. 
\section{Conclusion}
In this paper, we extended the application of DCT encoder to multi- and cross-lingual settings. Experimental results across different languages showed that similar to English using DCT outperform the vector averaging. We further presented a sentence-level-based approach for cross-lingual mapping without any additional training parameters. In this context, the DCT embedding is used to generate sentence representations, which are then used in the alignment process. 
Moreover, we have shown that incorporating structural information encoded in the lower-order coefficients yields significant performance gains compared to the AVG in sentence translation retrieval. 

\section*{Acknowledgments}
We thank Hanan Aldarmaki for providing us the in-domain FastText embeddings and for sharing many helpful insights. We would also like to thank 3 anonymous reviewers for their constructive feedback on this work.

\bibliographystyle{acl_natbib}
\bibliography{anthology,acl2021}

\begin{thebibliography}{24}
\expandafter\ifx\csname natexlab\endcsname\relax\def\natexlab#1{#1}\fi

\bibitem[{Ahmed et~al.(1974)Ahmed, Natarajan, and Rao}]{ahmed1974discrete}
Nasir Ahmed, T\_ Natarajan, and Kamisetty~R Rao. 1974.
\newblock Discrete cosine transform.
\newblock \emph{IEEE transactions on Computers}, 100(1):90--93.

\bibitem[{Aldarmaki and Diab(2019)}]{aldarmaki-diab-2019-context}
Hanan Aldarmaki and Mona Diab. 2019.
\newblock \href {https://doi.org/10.18653/v1/N19-1391} {Context-aware
  cross-lingual mapping}.
\newblock In \emph{Proceedings of the 2019 Conference of the North {A}merican
  Chapter of the Association for Computational Linguistics: Human Language
  Technologies, Volume 1 (Long and Short Papers)}, pages 3906--3911,
  Minneapolis, Minnesota. Association for Computational Linguistics.

\bibitem[{Almarwani et~al.(2019)Almarwani, Aldarmaki, and
  Diab}]{almarwani-etal-2019-efficient}
Nada Almarwani, Hanan Aldarmaki, and Mona Diab. 2019.
\newblock \href {https://doi.org/10.18653/v1/D19-1380} {Efficient sentence
  embedding using discrete cosine transform}.
\newblock In \emph{Proceedings of the 2019 Conference on Empirical Methods in
  Natural Language Processing and the 9th International Joint Conference on
  Natural Language Processing (EMNLP-IJCNLP)}, pages 3672--3678, Hong Kong,
  China. Association for Computational Linguistics.

\bibitem[{Bojanowski et~al.(2017)Bojanowski, Grave, Joulin, and
  Mikolov}]{bojanowski-etal-2017-enriching}
Piotr Bojanowski, Edouard Grave, Armand Joulin, and Tomas Mikolov. 2017.
\newblock \href {https://doi.org/10.1162/tacl_a_00051} {Enriching word vectors
  with subword information}.
\newblock \emph{Transactions of the Association for Computational Linguistics},
  5:135--146.

\bibitem[{Bojar et~al.(2013)Bojar, Buck, Callison-Burch, Federmann, Haddow,
  Koehn, Monz, Post, Soricut, and Specia}]{bojar-etal-2013-findings}
Ond{\v{r}}ej Bojar, Christian Buck, Chris Callison-Burch, Christian Federmann,
  Barry Haddow, Philipp Koehn, Christof Monz, Matt Post, Radu Soricut, and
  Lucia Specia. 2013.
\newblock \href {https://www.aclweb.org/anthology/W13-2201} {Findings of the
  2013 {W}orkshop on {S}tatistical {M}achine {T}ranslation}.
\newblock In \emph{Proceedings of the Eighth Workshop on Statistical Machine
  Translation}, pages 1--44, Sofia, Bulgaria. Association for Computational
  Linguistics.

\bibitem[{Chelba et~al.(2014)Chelba, Mikolov, Schuster, Ge, Brants, Koehn, and
  Robinson}]{DBLP:conf/interspeech/ChelbaMSGBKR14}
Ciprian Chelba, Tomas Mikolov, Mike Schuster, Qi~Ge, Thorsten Brants, Phillipp
  Koehn, and Tony Robinson. 2014.
\newblock \href
  {http://www.isca-speech.org/archive/interspeech\_2014/i14\_2635.html} {One
  billion word benchmark for measuring progress in statistical language
  modeling}.
\newblock In \emph{{INTERSPEECH} 2014, 15th Annual Conference of the
  International Speech Communication Association, Singapore, September 14-18,
  2014}, pages 2635--2639. {ISCA}.

\bibitem[{Conneau and Kiela(2018)}]{conneau-kiela-2018-senteval}
Alexis Conneau and Douwe Kiela. 2018.
\newblock \href {https://www.aclweb.org/anthology/L18-1269} {{S}ent{E}val: An
  evaluation toolkit for universal sentence representations}.
\newblock In \emph{Proceedings of the Eleventh International Conference on
  Language Resources and Evaluation ({LREC} 2018)}, Miyazaki, Japan. European
  Language Resources Association (ELRA).

\bibitem[{Conneau et~al.(2017)Conneau, Kiela, Schwenk, Barrault, and
  Bordes}]{conneau-etal-2017-supervised}
Alexis Conneau, Douwe Kiela, Holger Schwenk, Lo{\"\i}c Barrault, and Antoine
  Bordes. 2017.
\newblock \href {https://doi.org/10.18653/v1/D17-1070} {Supervised learning of
  universal sentence representations from natural language inference data}.
\newblock In \emph{Proceedings of the 2017 Conference on Empirical Methods in
  Natural Language Processing}, pages 670--680, Copenhagen, Denmark.
  Association for Computational Linguistics.

\bibitem[{Conneau et~al.(2018)Conneau, Kruszewski, Lample, Barrault, and
  Baroni}]{conneau-etal-2018-cram}
Alexis Conneau, German Kruszewski, Guillaume Lample, Lo{\"\i}c Barrault, and
  Marco Baroni. 2018.
\newblock \href {https://doi.org/10.18653/v1/P18-1198} {What you can cram into
  a single {\$}{\&}!{\#}* vector: Probing sentence embeddings for linguistic
  properties}.
\newblock In \emph{Proceedings of the 56th Annual Meeting of the Association
  for Computational Linguistics (Volume 1: Long Papers)}, pages 2126--2136,
  Melbourne, Australia. Association for Computational Linguistics.

\bibitem[{Devlin et~al.(2019)Devlin, Chang, Lee, and
  Toutanova}]{devlin-etal-2019-bert}
Jacob Devlin, Ming-Wei Chang, Kenton Lee, and Kristina Toutanova. 2019.
\newblock \href {https://doi.org/10.18653/v1/N19-1423} {{BERT}: Pre-training of
  deep bidirectional transformers for language understanding}.
\newblock In \emph{Proceedings of the 2019 Conference of the North {A}merican
  Chapter of the Association for Computational Linguistics: Human Language
  Technologies, Volume 1 (Long and Short Papers)}, pages 4171--4186,
  Minneapolis, Minnesota. Association for Computational Linguistics.

\bibitem[{Grave et~al.(2018)Grave, Bojanowski, Gupta, Joulin, and
  Mikolov}]{grave-etal-2018-learning}
Edouard Grave, Piotr Bojanowski, Prakhar Gupta, Armand Joulin, and Tomas
  Mikolov. 2018.
\newblock \href {https://www.aclweb.org/anthology/L18-1550} {Learning word
  vectors for 157 languages}.
\newblock In \emph{Proceedings of the Eleventh International Conference on
  Language Resources and Evaluation ({LREC} 2018)}, Miyazaki, Japan. European
  Language Resources Association (ELRA).

\bibitem[{Hill et~al.(2016)Hill, Cho, and
  Korhonen}]{hill-etal-2016-learning-distributed}
Felix Hill, Kyunghyun Cho, and Anna Korhonen. 2016.
\newblock \href {https://doi.org/10.18653/v1/N16-1162} {Learning distributed
  representations of sentences from unlabelled data}.
\newblock In \emph{Proceedings of the 2016 Conference of the North {A}merican
  Chapter of the Association for Computational Linguistics: Human Language
  Technologies}, pages 1367--1377, San Diego, California. Association for
  Computational Linguistics.

\bibitem[{Kayal and Tsatsaronis(2019)}]{kayal-tsatsaronis-2019-eigensent}
Subhradeep Kayal and George Tsatsaronis. 2019.
\newblock \href {https://doi.org/10.18653/v1/P19-1445} {{E}igen{S}ent: Spectral
  sentence embeddings using higher-order dynamic mode decomposition}.
\newblock In \emph{Proceedings of the 57th Annual Meeting of the Association
  for Computational Linguistics}, pages 4536--4546, Florence, Italy.
  Association for Computational Linguistics.

\bibitem[{Logeswaran and Lee(2018)}]{logeswaran2018efficient}
Lajanugen Logeswaran and Honglak Lee. 2018.
\newblock An efficient framework for learning sentence representations.

\bibitem[{Mikolov et~al.(2013)Mikolov, Sutskever, Chen, Corrado, and
  Dean}]{mikolov2013distributed}
Tomas Mikolov, Ilya Sutskever, Kai Chen, Greg~S Corrado, and Jeff Dean. 2013.
\newblock Distributed representations of words and phrases and their
  compositionality.
\newblock In \emph{Advances in neural information processing systems}, pages
  3111--3119.

\bibitem[{Ravishankar et~al.(2019)Ravishankar, {\O}vrelid, and
  Velldal}]{ravishankar-etal-2019-probing}
Vinit Ravishankar, Lilja {\O}vrelid, and Erik Velldal. 2019.
\newblock \href {https://doi.org/10.18653/v1/W19-4318} {Probing multilingual
  sentence representations with {X}-probe}.
\newblock In \emph{Proceedings of the 4th Workshop on Representation Learning
  for NLP (RepL4NLP-2019)}, pages 156--168, Florence, Italy. Association for
  Computational Linguistics.

\bibitem[{R{\"u}ckl{\'e} et~al.(2018)R{\"u}ckl{\'e}, Eger, Peyrard, and
  Gurevych}]{ruckle2018concatenated}
Andreas R{\"u}ckl{\'e}, Steffen Eger, Maxime Peyrard, and Iryna Gurevych. 2018.
\newblock Concatenated power mean word embeddings as universal cross-lingual
  sentence representations.
\newblock \emph{arXiv preprint arXiv:1803.01400}.

\bibitem[{Shao and Johnson(2008)}]{shao2008type}
Xuancheng Shao and Steven~G Johnson. 2008.
\newblock Type-ii/iii dct/dst algorithms with reduced number of arithmetic
  operations.
\newblock \emph{Signal Processing}, 88(6):1553--1564.

\bibitem[{Shu et~al.(2017)Shu, Wu, and Liu}]{shu2017study}
Xiao Shu, Xiaolin Wu, and Bolin Liu. 2017.
\newblock A study on quantization effects of dct based compression.
\newblock In \emph{2017 IEEE International Conference on Image Processing
  (ICIP)}, pages 3500--3504. IEEE.

\bibitem[{Toldova et~al.(2015)Toldova, Lyashevskaya, Bonch-Osmolovskaya, and
  Ionov}]{toldova2015evaluation}
S~Toldova, O~Lyashevskaya, A~Bonch-Osmolovskaya, and M~Ionov. 2015.
\newblock Evaluation for morphologically rich language: Russian nlp.
\newblock In \emph{Proceedings on the International Conference on Artificial
  Intelligence (ICAI)}, page 300. The Steering Committee of The World Congress
  in Computer Science, Computer~….

\bibitem[{Watson(1994)}]{watson1994image}
Andrew~B Watson. 1994.
\newblock Image compression using the discrete cosine transform.
\newblock \emph{Mathematica journal}, 4(1):81.

\bibitem[{Yang et~al.(2019)Yang, Zhu, and Chen}]{yang-etal-2019-parameter}
Ziyi Yang, Chenguang Zhu, and Weizhu Chen. 2019.
\newblock \href {https://doi.org/10.18653/v1/D19-1059} {Parameter-free sentence
  embedding via orthogonal basis}.
\newblock In \emph{Proceedings of the 2019 Conference on Empirical Methods in
  Natural Language Processing and the 9th International Joint Conference on
  Natural Language Processing (EMNLP-IJCNLP)}, pages 638--648, Hong Kong,
  China. Association for Computational Linguistics.

\bibitem[{Zhang et~al.(2018)Zhang, Xiong, and
  Su}]{zhang-etal-2018-accelerating}
Biao Zhang, Deyi Xiong, and Jinsong Su. 2018.
\newblock \href {https://doi.org/10.18653/v1/P18-1166} {Accelerating neural
  transformer via an average attention network}.
\newblock In \emph{Proceedings of the 56th Annual Meeting of the Association
  for Computational Linguistics (Volume 1: Long Papers)}, pages 1789--1798,
  Melbourne, Australia. Association for Computational Linguistics.

\bibitem[{Zhu and de~Melo(2020)}]{zhu-de-melo-2020-sentence}
Xunjie Zhu and Gerard de~Melo. 2020.
\newblock \href {https://www.aclweb.org/anthology/2020.coling-main.300}
  {Sentence analogies: Linguistic regularities in sentence embeddings}.
\newblock In \emph{Proceedings of the 28th International Conference on
  Computational Linguistics}, pages 3389--3400, Barcelona, Spain (Online).
  International Committee on Computational Linguistics.

\end{thebibliography}

\appendix
\section{Appendices}
\label{sec:appendix}
Table \ref{X-PROBE-Full-Results} shows the complete numerical results for the probing tasks on all languages.
\begin{table*}[!htbp]
\centering
\small
\begin{adjustbox}{width=0.80\textwidth}
\setlength\tabcolsep{4pt}
\begin{tabular}{cccccccc}
\hline
&Language&AVG&c[0]&c[0:1]&c[0:2]&c[0:3]&c[0:4]\\\hline
&EN&56.28&\textbf{89.03}&88.91&88.95&88.7&88.08\\
&ES&59.92&89.59&90.00&89.8&89.73&\textbf{90.05}\\
SentLen&FR&57.9&\textbf{93.72}&93.44&93.14&92.82&92.38\\
&DE&53.41&\textbf{88.81}&88.36&88.16&87.54&87.69\\
&RU&54.42&\textbf{89.66}&89.12&89.18&88.26&88.04\\\hline
&EN&26.97&\textbf{66.69}&64.55&62.49&60.39&59.08\\
&ES&25.4&\textbf{64.80}&62.18&60.62&58.76&57.64\\
WC&FR&27.14&\textbf{68.60}&66.13&64.71&62.8&61.04\\
&DE&29.33&\textbf{64.99}&64.52&63.93&63.12&61.54\\
&RU&36.33&\textbf{67.50}&65.58&64.69&62.69&61.32\\\hline
&EN&54.78&54.98&54.58&54.86&54.81&\textbf{55.58}\\
&ES&54.7&54.52&54.53&54.21&54.71&\textbf{55.77}\\
Bshift&FR&54.69&54.7&54.68&54.91&55.53&\textbf{56.50}\\
&DE&54.23&54.22&54.35&54.43&54.6&\textbf{56.46}\\
&RU&56.48&56.8&56.81&56.28&57.4&\textbf{58.51}\\\hline
&EN&41.34&45.18&48.64&49.84&49.44&\textbf{50.47}\\
&ES&42.9&48.53&52.29&53.34&\textbf{53.87}&53.54\\
TreeDepth&FR&41.06&47.68&50.05&51.65&\textbf{52.27}&52.15\\
&DE&37.06&41.97&45.14&47.33&\textbf{47.55}&47.36\\
&RU&35.27&39.21&40.76&\textbf{41.02}&40.65&40.51\\\hline
&EN&86.49&89.23&91.83&92.17&\textbf{92.26}&92.21\\
&ES&94.52&95.97&\textbf{96.68}&96.67&96.62&96.53\\
Tense&FR&91.96&94.06&95.7&95.96&\textbf{96.12}&95.99\\
&DE&94.13&94.71&95.82&\textbf{96.44}&96.28&95.92\\
&RU&86.07&86.39&90.28&\textbf{90.4}&90.16&90.38\\\hline
&EN&73.47&74.22&84.56&\textbf{87.20}&87.03&87.19\\
&ES&67.08&68.13&81.61&84.15&85.17&\textbf{85.77}\\
CoordInv&FR&71.06&71.12&85.97&88.03&89.21&\textbf{89.61}\\
&DE&74.25&74.33&89.99&92.52&93.45&\textbf{94.09}\\
&RU&60.33&60.77&79.95&83.13&84.03&\textbf{84.34}\\\hline
&EN&76.46&77.41&80.49&81.68&81.76&\textbf{82.31}\\
&ES&86.4&86.68&89.34&90.42&90.12&\textbf{90.84}\\
SubjNum&FR&88.48&88.62&91.05&92.23&92.72&\textbf{92.76}\\
&DE&75.94&75.78&78.79&78.9&79.25&\textbf{79.28}\\
&RU&70.47&70.44&72.31&72.81&73.12&\textbf{73.13}\\\hline
&EN&68.44&69.71&71.78&73.24&73.98&\textbf{74.93}\\
&ES&78.31&79.23&82.21&83.96&85.2&\textbf{85.7}\\
ObjNum&FR&77.47&78.5&83.74&85.82&86.92&\textbf{88.1}\\
&DE&68.38&68.74&69.88&70.41&71.14&\textbf{71.90}\\
&RU&63.9&63.79&65.33&65.32&\textbf{65.54}&65.11\\\hline
&EN&50.12&50.91&\textbf{51.72}&51.71&51.36&50.42\\
&ES&51.7&51.98&51.34&49.62&50.71&\textbf{53.07}\\
SOMO&FR&\textbf{50.7}&48.85&48.87&49.44&49.56&49.36\\
&DE&\textbf{50.57}&50.47&49.99&49.99&49.99&49.99\\
&RU&52.49&52.91&52.86&52.8&53.07&\textbf{53.13}\\\hline
\end{tabular}
\end{adjustbox}
\caption{DCT embeddings Performance per language compared to AVG. \textbf{EN}=English, \textbf{ES}=Spanish, \textbf{FR}=French, \textbf{DE}=German, and \textbf{RU}=Russian}
\label{X-PROBE-Full-Results}
\end{table*}

\end{document}